\documentclass[lettersize,journal]{IEEEtran}
\usepackage{amsmath,amsfonts}
\usepackage{algorithmic}
\usepackage[caption=false,font=normalsize,labelfont=sf,textfont=sf]{subfig}
\usepackage{textcomp}
\usepackage{stfloats}
\usepackage{url}
\usepackage{verbatim}
\usepackage{graphicx}
\usepackage{cite}

\usepackage{amsmath,amssymb,amsfonts}
\usepackage{multirow}
\usepackage[ruled,lined]{algorithm2e}

\begin{document}

\title{Autonomous Guidewire Navigation for Robot-assisted Endovascular Interventions: A Knowledge-Driven Visual Guidance Approach}

\author{Wentao Liu, Weijin Xu, Xiaochuan Li, Bowen Liang, Ziyang He, Mengke Zhu, Jingzhou Song, Huihua Yang and Qingsheng Lu
\thanks{This work was supported in part by the National Natural Science Foundation of China under Grant 62376038 and in part by the National Key R\&D Program of China under Grant 2018AAA0102600. (Corresponding author:   Huihua yang and Qingsheng Lu)

Wentao Liu is with the Dynamic Products Department, Wandong Medical Technology Co., Ltd, Beijing 100015, China (e-mail: liuwt61@midea.com).

Weijin Xu is with the Department of Cyberspace Defense, Beijing Police College, Beijing 102202, China (e-mail: xuweijin@bjpc.edu.cn)

Xiaochuan Li, Ziyang He, Jingzhou Song and Huihua Yang are with the School of Intelligent Engineering and Automation, Beijing University of Posts and Telecommunications, Beijing 100876, China (e-mail: 939972028@bupt.edu.cn; heziyang@bupt.edu.cn; sjz2008@bupt.edu.cn; yhh@bupt.edu.cn).

Bowen Liang and Qingsheng Lu are with the Department of Vascular Surgery, Shanghai Changhai Hospital, Naval Medical University, Shanghai 200433, China
(e-mail: liangbowen\_vas@163.com;
luqs@newvascular.cn)

Mengke Zhu is with the State Key Laboratory of Space-Ground Integrated Information Technology, Beijing Institute of Satellite Information Engineering, Beijing 100086, China (e-mail: zmk19@tsinghua.org.cn).

}

}

\markboth{Journal of \LaTeX\ Class Files,~Vol.~14, No.~8, August~2021}%
{Shell \MakeLowercase{\textit{et al.}}: A Sample Article Using IEEEtran.cls for IEEE Journals}


\maketitle
\begin{abstract}
Autonomous robots for endovascular interventions hold significant potential to enhance procedural safety and reliability by navigating guidewires with precision, minimizing human error, and reducing surgical time. However, existing methods of guidewire navigation rely on manual demonstration data and have a suboptimal success rate. In this work, we propose a knowledge-driven visual guidance (KVG) method that leverages available visual information from interventional imaging to facilitate guidewire navigation. Our approach integrates image segmentation and detection techniques to extract surgical knowledge, including vascular maps and guidewire positions. We introduce BDA-star, a novel path planning algorithm with boundary distance constraints, to optimize trajectory planning for guidewire navigation. To validate the method, we developed the KVD-Reinforcement Learning environment, where observations consist of real-time guidewire feeding images highlighting the guidewire tip position and the planned path. We proposed a reward function based on the distances from both the guidewire tip to the planned path and the target to evaluate the agent's actions.Additionally, to address stability issues and slow convergence rates associated with direct learning from raw pixels, we incorporated a pre-trained convolutional neural network into the policy network for feature extraction. Experiments conducted on the aortic simulation autonomous guidewire navigation platform demonstrated that the proposed method, targeting the left subclavian artery, left carotid artery and the brachiocephalic artery, achieved a 100\% guidewire navigation success rate, along with reduced movement and retraction distances and trajectories tend to the center of the vessels.
\end{abstract}

\begin{IEEEkeywords}
Robot-Assisted endovascular interventions, autonomous guidewire navigation, visual guidance
\end{IEEEkeywords}

\section{Introduction}
\IEEEPARstart{C}{ardiovascular} diseases (CVDs) remain the leading cause of mortality worldwide~\cite{2022global}. Endovascular intervention (EI), a sophisticated technique within the field of endovascular therapy, is primarily employed for the diagnosis and treatment of cardiovascular diseases. Known for its minimal invasiveness, rapid recovery, and low complication rates, EI is extensively applied in managing various cardiovascular conditions, including aortic disease, peripheral arterial disease, and stroke. To mitigate the persistent risk of radiation exposure faced by surgeons during fluoroscopic procedures, several robotic systems based on leader-follower teleoperation architecture have been developed. These systems enable surgeons to control a leader device, whose inputs are mapped to a follower robot that executes the corresponding actions. This allows surgeons to perform procedures remotely from a radiation-shielded environment, significantly enhancing safety. Recent advancements in robotic platforms for endovascular interventions have demonstrated their potential to assist in the successful completion of procedures. However, unlike the direct manipulation of surgical instruments such as guidewires and catheters, operating surgical robots relies heavily on direct physician control. This leads to longer surgical durations and a steeper learning curve. Robot-assisted autonomous guidewire navigation (RAGN) has emerged as a promising solution to improve surgical efficiency without relying on expert experience~\cite{cathsim}.

Recently, RAGN has undergone significant advancements, leading to several notable innovations. These systems utilize deep learning techniques, such as convolutional neural networks (CNNs) and generative adversarial networks (GANs), to enable surgical status prediction~\cite{zhao2019cnn}, catheter tip tracking~\cite{ma2020dynamic}, vessel segmentation~\cite{dias,su2022cave}, and automated catheterization~\cite{chi2020collaborative}. Furthermore, deep reinforcement learning (RL) techniques have demonstrated the ability to model large, continuous state spaces, thereby facilitating the automation of complex EI tasks. These techniques can learn relevant behavioral patterns from human demonstrations, which encapsulate surgical operational experience, by iteratively updating strategies through environmental interactions~\cite{chi2018trajectory,chi2020collaborative,casog}. Currently, autonomous guidewire navigation has been successfully implemented in virtual surgical environments~\cite{jianu2024autonomous} and on two-dimensional planar vascular platforms~\cite{casog,kweon2021deep}. However, most vision-guided methods exhibit low navigation success rates, and their reliance on manual demonstrations limits scalability~\cite{casog}. Additionally, electromagnetic-based guidewire positioning has proven impractical for real-world surgical scenarios~\cite{hwang2020review}.


During endovascular interventions, surgeons monitor the guidewire's real-time position on DSA images and manually guide the tip to the lesion using the angiogram as a roadmap. The extraction and quantification of expert knowledge, including the guidewire position and vascular roadmap on DSA images, are critical to the success of RAGN. In this paper, we propose a knowledge-driven visual guidance (KVD) method for RAGN that operates independently of expert demonstrations. The main contributions of this work are as follows:

\begin{enumerate}
\item We utilize FR-UNet~\cite{frunet} and YOLOv8~\cite{yolov8_ultralytics} to facilitate the segmentation of vascular models and the localization of the guidewire tip within the images captured by the RAGN platform. We introduce the BDA-star path planning algorithm, which simultaneously optimizes trajectory distance and boundary proximity.

\item We employ a pre-trained model for representation extraction, leveraging its prior knowledge in recognizing generic patterns and features within images. We established an RL environment for guidewire navigation, which incorporating observations that integrate multi-image information and a reward function specifically designed for path navigation. 

\item Experiments conducted on the aortic simulation KVD platform demonstrate the efficacy of the proposed method. Targeting the Left Subclavian Artery (LSA), Left Carotid Artery (LCA), and Brachiocephalic Artery (BCA), the method achieved a 100\% guidewire navigation success rate. Additionally, this approach resulted in reduced movement and retraction distances, with trajectories consistently converging toward the central axis of the vessels.

\end{enumerate}




\section{Related work}





Currently, numerous studies on autonomous guidewire navigation have achieved significant progress at the experimental stage~\cite{yang2022guidewire}. Rafii-Tari et al.~\cite{rafii2013learning} proposed a learning-from-demonstration framework for robot-assisted catheter insertion, where motion models derived from expert demonstrations are replicated by a robotic catheter driver to assist less experienced surgeons. This approach offers new possibilities for surgeon-robot collaboration and semi-automation, reducing the cognitive workload of vascular surgeons. Chi et al.~\cite{chi2018trajectory} introduced a trajectory optimization method for robot-assisted catheter insertion, combining path-integral reinforcement learning with dynamic movement primitives to semi-automate the task. This method reduces path length, contact force, and the risks of vascular injury, embolism, and stroke. Furthermore, Chi et al.~\cite{chi2020collaborative} developed a semi-automated strategy for robot-assisted endovascular catheter insertion using imitation learning. In this approach, latent skill patterns from expert demonstrations are extracted, and a model-free generative adversarial imitation learning method is employed to automate arterial catheterization. Compared to other methods, learning-from-demonstration and imitation learning maximize the acquisition of surgeons' behavioral patterns and operational habits. However, the action models learned may vary across different surgeons, making it challenging to generalize these models to other interventional procedures.



\begin{figure}[ht]
\centering
\includegraphics[scale=0.31]{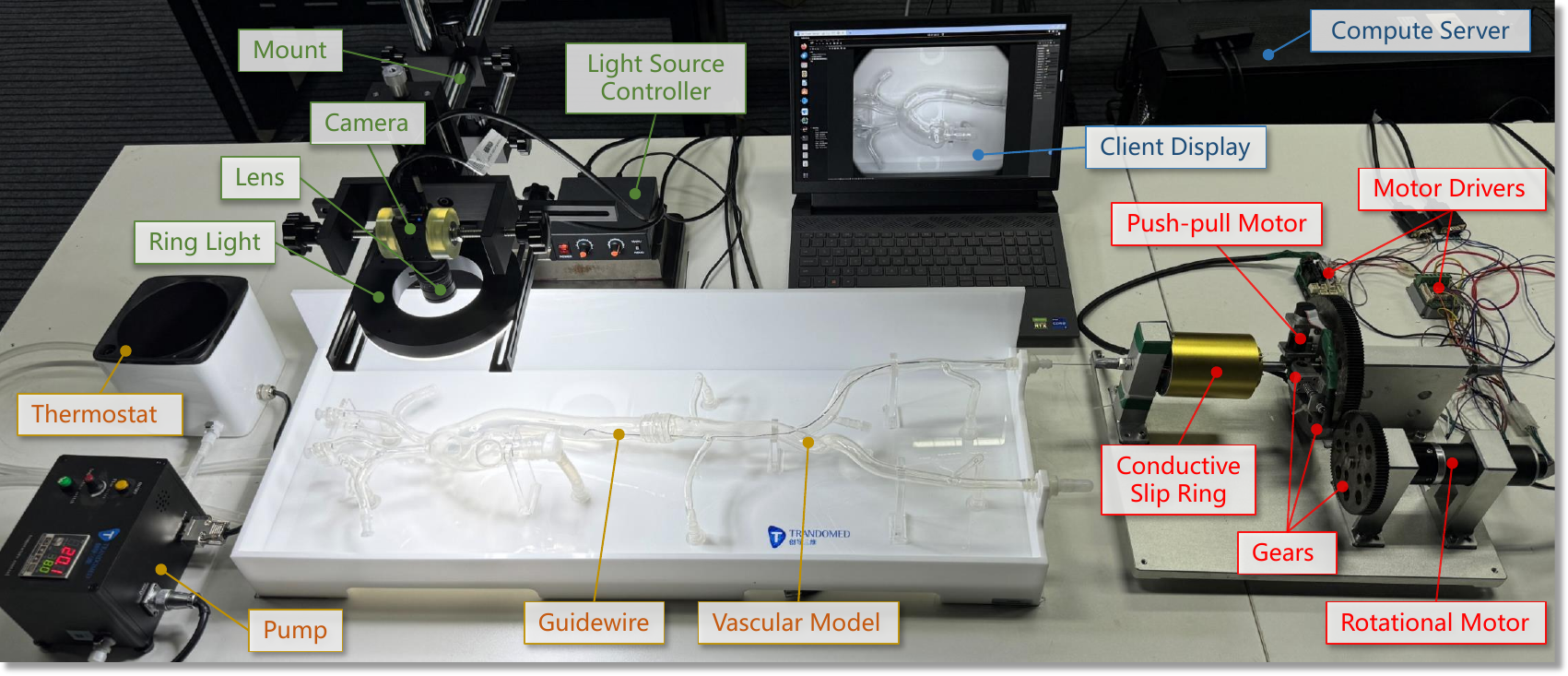}
\caption{Image-guided autonomous guidewire navigation platform.} \label{fig:platform}
\end{figure}

RL enables autonomous optimization through continuous interaction with the environment, allowing for more precise and efficient navigation in complex and dynamic vascular conditions. Behr et al.~\cite{behr2019deep,karstensen2020autonomous} utilized Deep Q-Networks (DQN) and Deep Deterministic Policy Gradients (DDPG) to train and test autonomous guidewire navigation in simulated vascular models. By incorporating experience replay and human demonstrations, they accelerated the training process, enabling the guidewire to autonomously reach random targets from random starting points. Kweon et al.~\cite{kweon2021deep} proposed an autonomous guidewire navigation system using the Rainbow algorithm, which employs staged learning, human demonstrations, transfer learning, and weight initialization to expedite training. They defined the "state" as the area near the guidewire tip and introduced sub-goals to mitigate sparse rewards and ensure target reachability. Li et al.~\cite{li2023casog} introduced an offline reinforcement learning algorithm (CASOG) designed to learn efficient manipulation skills for vascular robotic systems. This approach mitigates distributional shift and overfitting by conservatively estimating the Q-function and smoothing convolutional gradients, while prioritizing transitions with large temporal difference errors to enhance the learning of complex operations. In summary, while these technologies have made significant progress in autonomous guidewire navigation, they remain heavily reliant on demonstration data and have primarily been validated on 2D vascular model platforms. This represents a substantial gap compared to the 3D guidewire navigation environment encountered in clinical practice.

\section{Materials and Methods}

\subsection{Autonomous Guidewire Navigation Platform}
As shown in Fig.~\ref{fig:platform}, we have established a RAGN platform consisting of a machine vision system, an anthropomorphic vessel model, a guidewire feeding robot, and a compute server. The vessel model includes the femoral artery, iliac artery, Type-I aortic arch, and carotid arteries, and is constructed from a transparent silicone-based material (Ningbo Chuangdao 3D Medical Technology Co., Ltd., China). The guidewire used for the task is a 0.035-inch nitinol guidewire (RF*GA35153M, Terumo, Japan), known for its exceptional flexibility and maneuverability. The machine vision system comprises a camera (MV-CS050-10GM, Hikvision, China), a lens (MVL-HF0624M-10MP, Hikvision, China), a ring light, and a mount. It is vertically positioned directly above the aortic arch to meet the requirements of guidewire navigation tasks. The captured images cover the vasculature from the thoracic aorta to the LSA, LCA, or BCA. To meet the computational demands for tasks such as model segmentation, guidewire tip positioning, and autonomous guidewire delivery training, the compute server is equipped with dual Intel(R) Xeon(R) Silver 4210R CPUs operating at 2.40 GHz each, a GeForce RTX 3090 GPU, and 128 GB of RAM.


The right side of Fig.~\ref{fig:platform} presents our custom-designed guidewire feeding robot, which is capable of executing push-pull and rotational actions on the guidewire for conducting IAGN training through RL. The push-pull motor (A-max 19, Maxon, Switzerland) is employed to drive the rotation of the active friction wheel in the friction wheel assembly. This mechanism utilizes frictional force to propel the guidewire, which is tightly gripped by the friction wheels, to move forward or backward. To avoid the limitations of wire entanglement and eliminate the need for reset operations, a conductive slip ring (MT1269-P0205-S11-VC, Moflon, China) is used to connect the push-pull motor and its driver (EPOS4 Compact 24/1.5 CAN, Maxon, Switzerland). The rotational motor (ECG40, Vichuan, China), in conjunction with its driver (SOL-WHI20/48E05, Elmo, Israel), drives the passive gear and the components attached to it to rotate collectively, enabling the rotation of the guidewire that is secured by the friction wheels.

The right side of Fig.~\ref{fig:platform} showcases our custom-designed guidewire feeding robot, which is capable of executing push-pull and rotational actions on the guidewire to facilitate KVD training through RL. The push-pull motor (A-max 19, Maxon, Switzerland) drives the rotation of the active friction wheel in the friction wheel assembly. This mechanism utilizes frictional force to propel the guidewire, which is tightly gripped by the friction wheels, enabling forward or backward movement. To avoid wire entanglement and eliminate the need for reset operations, a conductive slip ring (MT1269-P0205-S11-VC, Moflon, China) is used to connect the push-pull motor and its driver (EPOS4 Compact 24/1.5 CAN, Maxon, Switzerland). The rotational motor (ECG40, Vichuan, China), in conjunction with its driver (SOL-WHI20/48E05, Elmo, Israel), drives the passive gear and its attached components to rotate collectively, enabling the rotation of the guidewire secured by the friction wheels.

\begin{figure}[t]
\centering
\includegraphics[scale=0.32]{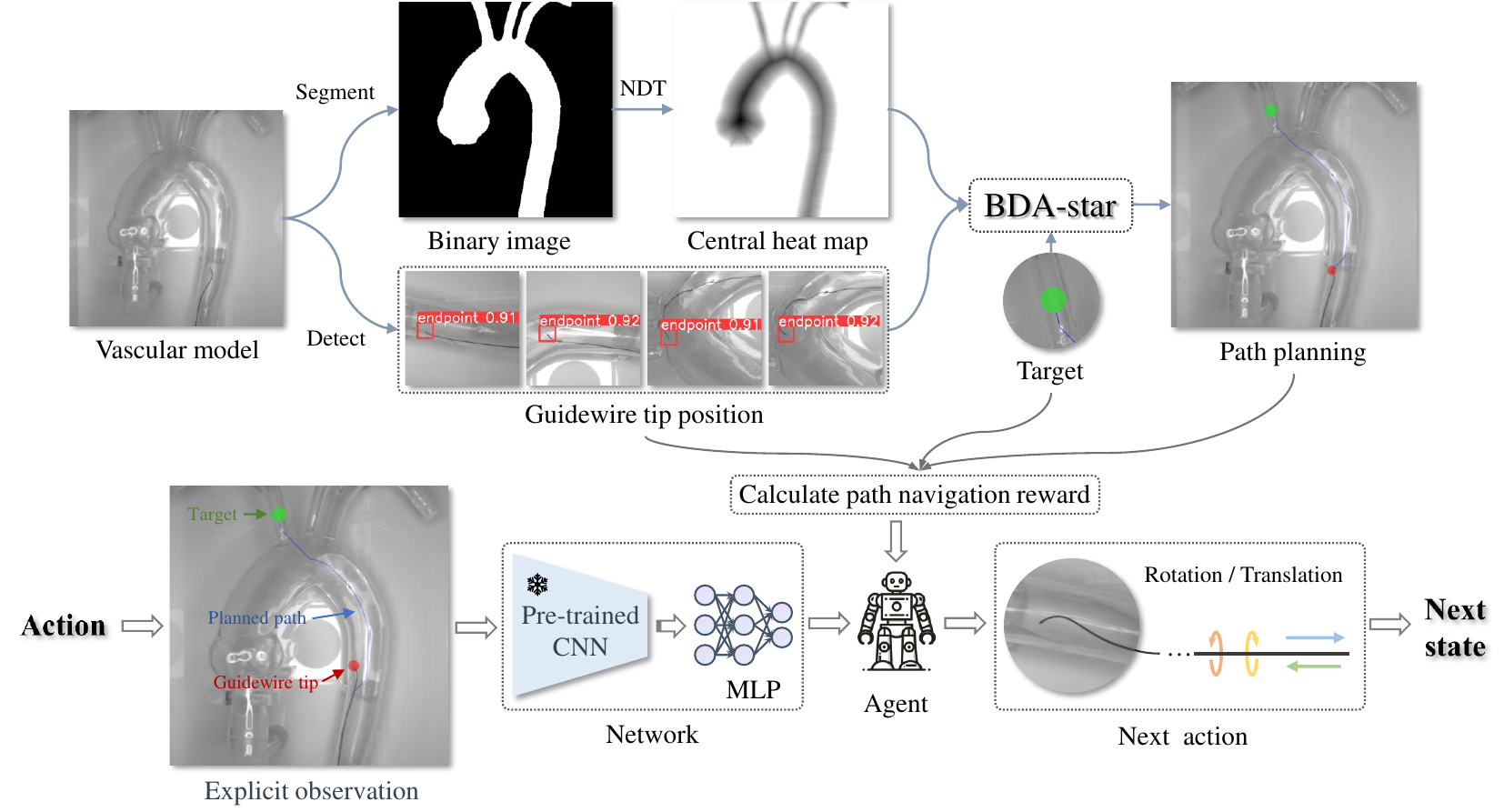}
\caption{Overview of the proposed image-guided autonomous navigation framework for endovascular interventions (NDT: Normalized Distance Transform).} \label{fig:method}
\end{figure}

\subsection{Trajectory Navigation}
The path planning of the guidewire and the accurate positioning of its tip are fundamental to achieving effective trajectory navigation during guidewire feeding. The following sections introduce the methods proposed for addressing them.


\subsubsection{Path Planning} 
In the guidewire navigation task, path planning is defined as determining the optimal path within the vessel between the guidewire tip and the target point (vessel branch or lesion). The most prevalent path planning algorithms are A-star~\cite{A*} and its derivatives, which are based on heuristic pathfinding and graph traversal techniques. However, to apply these algorithms to EI, it is first necessary to segment the vessels to define their boundaries and thus confine the search space within the vessels. We collected 10 images of the vessel model and performed pixel-level annotations. These images were then used to train the 2D segmentation network, FR-UNet~\cite{frunet}, specifically for vessel segmentation. The trained model is subsequently employed for the real-time segmentation of vascular models captured by the camera. 



To minimize the path distance while directing the trajectory towards the center of the vessel and to reduce the likelihood of collisions between the guidewire tip and the vessel walls as much as possible, we propose BDA-star. This algorithm incorporates a boundary distance measure for each node into the cost function, enabling a balance between the costs of path length and boundary proximity. Specifically, we utilize the segmented vascular image to generate a boundary distance heatmap. Initially, a distance transformation is conducted on the segmented map to produce a heatmap representing the distance from each vascular pixel to the nearest zero pixel. Subsequently, a convolution kernel created based on the heatmap's maximum value is used to convolve the segmented map. Finally, the heatmap is normalized by dividing it with the convolved image. Let $\mathbf{I}$  be the segmented map, the normalized heatmap $\mathbf{H}$ is given by:
\begin{equation}
\mathbf{H} = \frac{D(\mathbf{I})}{\mathbf{I} \ast K_{\max(D(\mathbf{I}))}},
\end{equation}
where $D(\mathbf{I})$ is the euclidean distance transform of $V$, $\ast$ represents the convolution operation, and $K_{\max(D(\mathbf{I}))}$ is the convolution kernel generated from the maximum value of $D(\mathbf{I})$.

We define each pixel within the segmentation image as a node, represented by $n$. Among the paths expanded from each node, the path in the BDA-star algorithm is selected by minimizing the cost function:
\begin{equation}
f(n) = \underbrace{g(n) + h(n)}_{A-star} + \omega \mathbf{H}(n),
\end{equation}
where $g(n)$ is the cost of the path from the start node to node $n$, $h(n)$ is a heuristic function that estimates the cost from node $g(n)$ to the goal. In KVD task, the heuristic function is the Euclidean distance. $\mathbf{H}(n)$ represents the distance between the node $n$ and the nearest boundary node. $W$ represents the weight of the boundary distance term. The greater the value of $W$ is, the more closely the path aligns with the center of the vessel.

\subsubsection{Guidewire Tip Localization}
 Currently, the prevalent methods for locating the tip of guidewires primarily rely on image segmentation~\cite{Sendpoint,Sendpoint2}, with a minority employing a two-phase strategy that combines initial segmentation followed by detection~\cite{TSendpoint}. To maximize the efficiency of localization, we directly employ object detection for the positioning of the guidewire tip. For this purpose, we recorded a 10-minute video of manual guidewire feeding, annotating these images with rectangular bounding boxes centered around the guidewire tip. Subsequently, we train this dataset using the one-stage detection framework YOLOv8~\cite{yolov8_ultralytics}, resulting in a highly effective model that is utilized for real-time positioning of the guidewire tip during autonomous guidewire navigation.

\subsection{Reinforcement Learning for Autonomous Guidewire Navigation} 
\subsubsection{Problem Definition} 
 We conceptualize KVD as an episodic partially observable markov decision process. The agent, embodied by the guidewire, navigates through an environment $E$, represented by the vessel model. At each discrete time step $t$, the agent receives an observation $s_t$, based on which it selects an action $a_t(s_t)$. This action leads to a reward $r_t(s_t,a_t)$ and transitions the agent to a new state $s_{t+1}$. The episodic process culminates when the agent attains its goal position within the vessel, denoted as $g$ in the goal state set $G$.




\subsubsection{Explicit Observations}
In the KVD platform, the camera deployed above the aorta captures real-time guidewire feeding images, akin to 2D Digital Subtraction Angiography (DSA) seen in actual surgical scenarios~\cite{dias}. The surgeon observes and analyzes these images to perform EI procedures. Likewise, in KVD task, we consider only the guidewire feeding images as the observations for the agent. Specifically, we fuse the pre-planned path, target, and the real-time positioning of the guidewire tip with live imaging. This explicit observation capability is designed to focus the learning interest of RL task. This enhanced observation explicitly articulates useful state information, which aids in accelerating policy learning and improving the robustness of the algorithm. Given the original image $I$, the initial guidewire tip position 
$p_g$, the target position $p_t$, and the set of path points $S$, the explicit observation is generated as follows:

\resizebox{0.48\textwidth}{!}{$
I'(x, y) =
\begin{cases} 
(1 - \alpha_1) I(x, y) + \alpha_1 C_1, & \text{if } \| (x, y) - P_g \| \leq R_1 \\
(1 - \alpha_2) I(x, y) + \alpha_2 C_2, & \text{if } \| (x, y) - P_t \| \leq R_2 \\
(1 - \alpha_3) I(x, y) + \alpha_3 C_3, & \text{if } \exists P_i \in S, \| (x, y) - P_i \| \leq R_3 \\
I(x, y), & \text{otherwise}
\end{cases},
$}
where, $R_1$, $R_2$ and $R_3$ represent the radii around the starting point, ending point, and path points, respectively. $C_1$, $C_2$ and $C_3$ represent the color values. $\alpha_1$, $\alpha_2$ and $\alpha_3$, which are in the range [0, 1], denote the blending factors for mixing the original image and the color values.

\subsubsection{Actions} 


The actions are represented by a vector $a_t \in \mathbb{R}^{2}$, which are associated respectively with the push-pull motor that drives the guidewire's translational movement and the motor responsible for the guidewire's rotation. We have set the maximum translational distance $S$ and the maximum rotation angle $R$ for each action step of the guidewire, thereby establishing the action spaces as $[-S,S]$ for translation and $[-R,R]$ for rotation, respectively. The computing server sends motion commands containing both translation and rotation information to the interventional surgical robot via serial communication. In the experiment, the push-pull motor operates in speed control mode, maintaining a constant rotational speed during execution. Given the push-pull motor's rotational speed $rpm_1$, The time required for the push-pull motor to rotate when the guidewire moves a distance of $S$ millimeters can be calculated as::

\begin{equation}
  T_P= \frac{60000\cdot S}{2\pi \cdot rpm \cdot d \cdot (r+\epsilon)},
\end{equation}
where, $v$ represents the guidewire's movement speed, measured in millimeters per millisecond, and  $d$ denotes the motor's reduction ratio. $r$ is the radius of the friction wheel, also measured in millimeters. When calculating the guidewire's movement speed using angular velocity and rotation radius, the guidewire's own radius should also be factored in. However, the friction wheel used in this study has some elasticity and a grooved structure. In practical applications, any discrepancies between the preset and actual movement distances are corrected by adjusting for an error term $\epsilon$, ensuring accurate translation of the guidewire.

Similarly, for the motor driving the guidewire's rotation, the given rotation angle $\theta$ must be converted into the corresponding rotation time (ms). The calculation formula is as follows:
\begin{equation}         
 T_R = \frac{60000\cdot\theta}{360\cdot rpm \cdot d \cdot c},
\end{equation}
where, $c$ represents the diameter ratio between the driving gear and the driven gear.




\subsubsection{Path Navigation Reward Function}
In the context of an KVD system, the reward function is designed to aid the system in guiding the guidewire tip towards a target point, typically within the vascular network of a patient during a medical procedure such as angiography or stent placement. The reward function is composed of three parts, reflecting different stages and strategies within the navigation process. 

\begin{enumerate}
    \item Success Reward: This is a fixed reward $R_{\text{s}}$ given when the guidewire tip is within a threshold distance $\delta$ from the target point. This condition reflects the successful completion of the navigation task. 
    \item Boundary Penalty Reward: We set movement range limits $[d_f, -d_b]$ to restrict the guidewire's displacement and assist in training. If the cumulative movement exceeds the set range, a boundary penalty reward $R_{\text{bp}}$ is applied and training is terminated.
    \item Continuous Reward: When the guidewire tip is not within the threshold distance, the reward function is designed to encourage progress towards the target point and adherence to a pre-planned path, while penalizing deviations from the optimal trajectory and promoting smoothness of the trajectory.
\end{enumerate}

The formula for the reward function $r(t)$ at time $t$ is defined as:
\resizebox{0.48\textwidth}{!}{$
r(t) = \begin{cases} 
R_{\text{s}}, & \text{if } \|x_t - g\|_2 \leq \delta \\
R_{\text{bp}}, & \text{else if } \sum\limits_{i=1}^{t} s_i \notin [d_f, -d_b] \\
-\left( e^{\omega_1 \min\limits_{k=1}^N \|x_t - p_k\|_2} + \omega_2 \sum\limits_{k=j}^{N-1} \|p_{k+1} - p_k\|_2 \right), & \text{otherwise}
\end{cases}
$}
where $x_t$ denotes the position of the guidewire tip at time $t$, $g$ represents the target point's position. The norm $\|\cdot\|_2$ denotes the euclidean distance. $s_i$ represents the distance moved by the guidewire at time step $i$. $\omega_1$ and $\omega_2$ are weighting factors that adjust the importance of the two terms in the continuous reward. $N$ is the number of points on the planned path. $p_k$ represents the $k$-th point on the planned path. The $1$-st and $N$-th points serve as the starting and ending points, respectively. $j$ is the index of the point on the pre-planned path that is closest to the guidewire tip, i.e., $\arg \min_{k=1,\ldots,N} \|x_t - p_k\|_2$.



\subsubsection{Network Architectures} 

We employ a pre-trained CNN to extract features from input images i.e., observations. During training, the weights of this CNN model are frozen, and the resulting feature vectors are fed into a Multi-Layer Perceptron (MLP) for policy training. This approach leverages the pre-trained model's prior knowledge in recognizing generic patterns and features within images. It offers a more stable and meaningful input space while effectively reducing data dimensionality. 

\begin{algorithm}
\small
    \caption{KVD}\label{alg1}
    \KwIn{Max translation distance per step $D$; Max rotation angle per step $R$; Max episodes $E$; Max time steps $T$; Threshold distance $\delta$; Movement range limits $[d_f, -d_b]$; Target position $g$.

    }
    \KwResult{Guidewire navigates to target vascular position autonomously}

    Segment vessels, locate guidewire tip and plan path.\\
    \If{Testing mode}{
    Load pre-trained model weights}
    \ForEach{E}{
    Reset environment.
            \For{each t in T } {
            Select action translation $d$ and rotation $r$ via policy network;\\
                Convert the action into motor parameters and send them to the driver for execution;\\
                Locate new guidewire tip from current image, update position $x_t$;\\
                Constructs a explicit observation of the current state;\\
                Calculate new reward and the accumulated movement distance S;\\
                \If{Training mode}{
                Store [state, action, reward, next state] in experience replay buffer;}
                \If{$\|x_t - g\|_2 \leq \delta$ or $S \notin [d_f, -d_b]$ }{
                \KwRet{break}
                 }
      
        }
  \If{Training mode}{
  Update agent from experience replay buffer}
    Retract the guidewire to the initial position according to S.
   }
    \If{Training mode}{
                Save model weights}
    
    \end{algorithm}
\section{Experiments}

\subsection{Implementation Details}

We employed the Soft Actor-Critic (SAC) algorithm for RL training in autonomous guidewire navigation. At its core, SAC is an off-policy, actor-critic framework that ingeniously incorporates the principle of entropy maximization to foster exploration, thereby enhancing the robustness and efficiency of policy learning. The SAC algorithm emerges as a cutting-edge approach within the reinforcement learning domain, particularly tailored for environments characterized by continuous action spaces. 

The experiments were conducted using PyTorch, and for the implementation of the SAC, we employed stable baselines~\cite{stable-baselines3}. We targeted the LSA, LCA, and BCA for guidewire navigation, with training time steps of 150,000, 300,000, and 300,000, respectively. The maximum number of steps allowed for each episode was set to 50. The maximum values for actions $S$ and $R$ are set to 20 mm and 90°. We utilize a pre-trained ResNet50~\cite{he2016deep} for feature extraction. In the reward function, the reward $R_{success}$ for successful navigation is set to 50, and the threshold distance $\delta$ is set to 40. The parameters $\omega_1$ and $\omega_2$ are set to 0.005 and 0.01, respectively. In DBA-star, the weight of the boundary distance term is set to 2. We conducted 20 tests each for the LSA, LCA, and BCA tasks, calculating the success rates, rewards and lengths of episodes, movement distances, boundary distances, and retracement distances to evaluate the performance of the KVD algorithm. We set a baseline using original images as observations, with the reward function's second term being the negative euclidean distance between the target point and the guidewire tip's real-time position~\cite{cathsim}, employing stable baselines's CnnPolicy as the policy network. On the baseline, Pre-trained CNNs (\textit{PC}), Path Navigation Rewards (\textit{PNR}), and Explicit Observations (\textit{EO}) were incrementally integrated to evaluate the performance of each part as well as their collective efficacy. The method of calculating Path Navigation Rewards along the path differs from that of Euclidean distance, resulting in lower reward values, making them incomparable.

\subsection{Quantitative Results}
\begin{figure}[ht]
\centering
\includegraphics[scale=0.28]{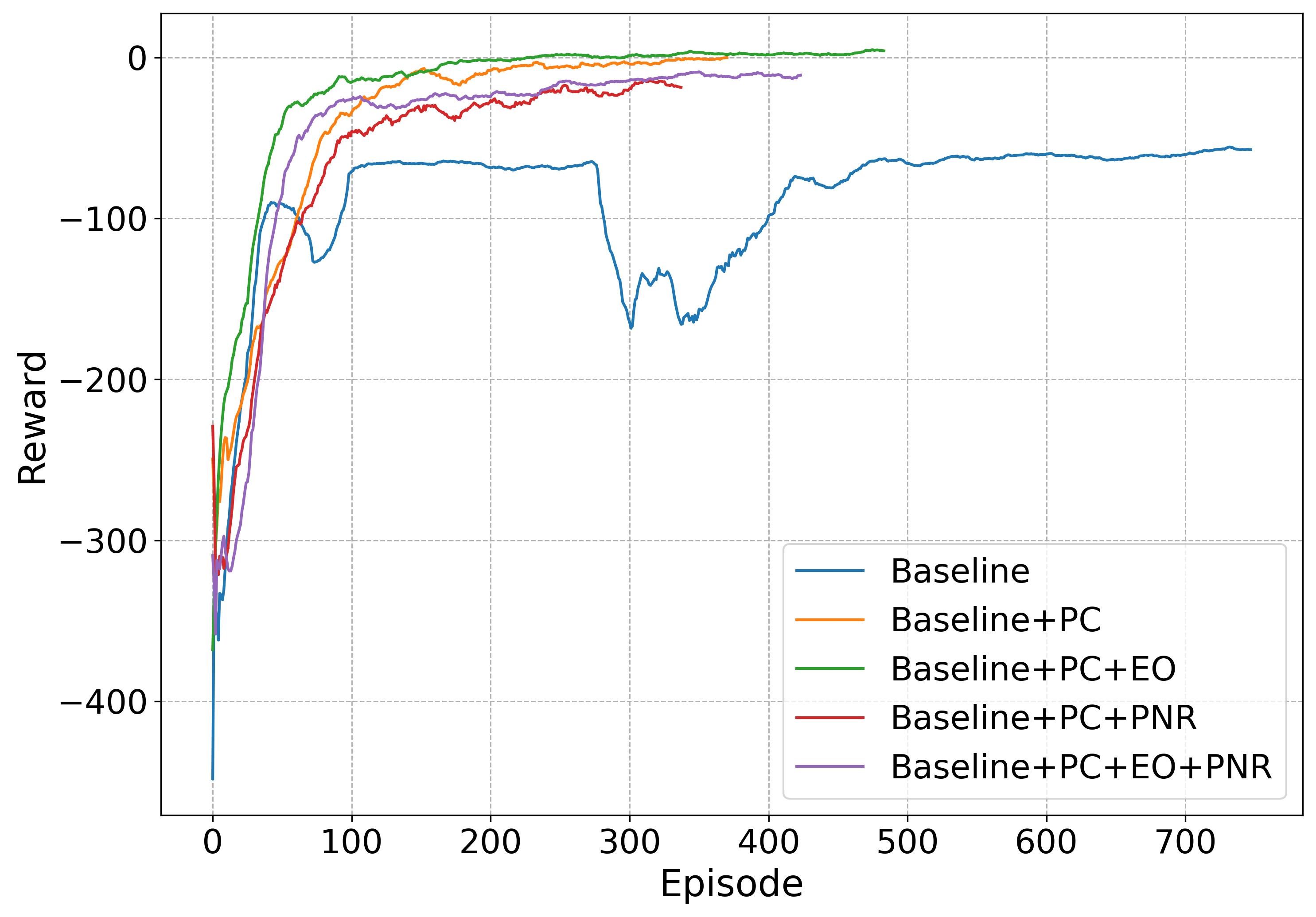}
\caption{The curve of the mean reward per episode during training in the BCA navigation task} 
\label{fig:train_reward}
\end{figure}

\begin{figure}[ht]
\centering
\includegraphics[scale=0.28]{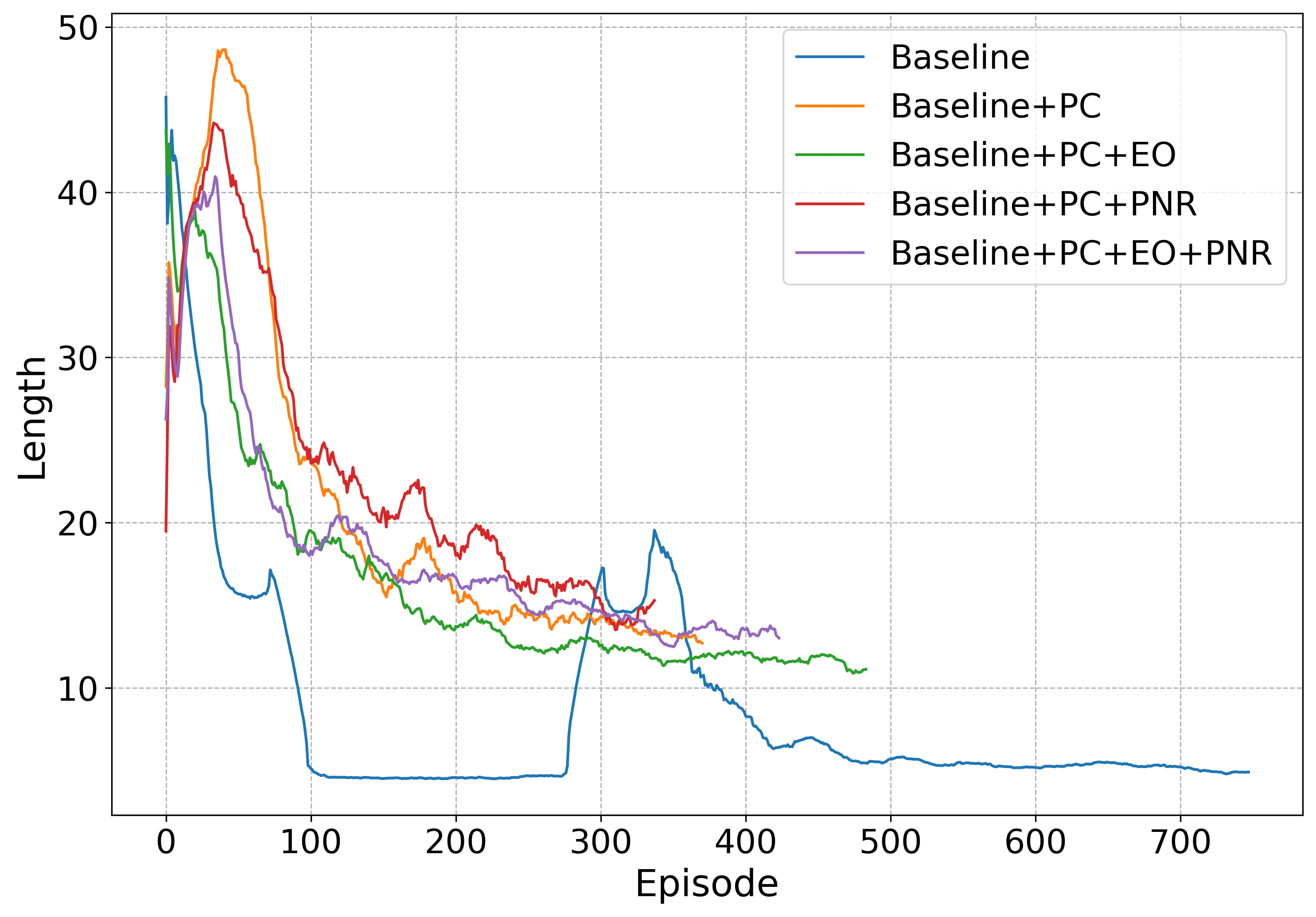}
\caption{The curve of the lengths per episode during training in the BCA navigation task} 
\label{fig:train_length}
\end{figure}

Fig.~\ref{fig:train_reward} and Fig.~\ref{fig:train_length} illustrate the average reward and episode lengths during training in the BCA navigation task. The low average reward of the baseline indicates a lower success rate for BCA guidewire navigation. Most episodes ended with lengths under 5, suggesting that backward movements of the guidewire frequently caused the training episodes to terminate. The strategy of using pre-trained CNNs for feature extraction, i.e., \textit{baseline+PC}, significantly increased the average reward, indicating a higher success rate in guidewire navigation. Among these methods, \textit{baseline+PC+EO} achieved the highest average reward and the shortest episode lengths, excluding the baseline. The method of calculating path navigation rewards along the path differs from that of euclidean distance, resulting in lower reward values, making them incomparable. Since this reward function tends to guide the guidewire toward the center of the vessel, it restricts the range of movement, leading to relatively longer episode lengths.

\begin{table*}[ht]
\centering
\caption{Experiment results. $\dagger$ indicated that the continuous reward uses negative Euclidean distance, resulting in higher values than PNR.  $\ddagger$ means the policy of \textit{baseline} is restricted to forward motion and is not considered in the comparative evaluation.}
\label{tab:vs}
\renewcommand\arraystretch{1.5}
\tabcolsep3pt                 
\scalebox{0.9}{
\begin{tabular}{cccccccccc}
\hline
\multirow{2}{*}{Tasks} & \multicolumn{3}{c}{Methods}                      & \multicolumn{6}{c}{Metrics}                                                                                              \\ \cline{2-10} 
                       & \textit{PC} & \textit{EO}  & \textit{PNR}   & SR \% $\uparrow$     & Episode reward $\uparrow$  & Episode length  $\downarrow$                & Movement distances (mm) $\downarrow$                    & Boundary distances (px)  $\uparrow$   & Retracement distances (mm) $\downarrow$      \\ \hline
\multirow{5}{*}{LSA}   & \multicolumn{3}{c}{Baseline}                     & 65           & -125.3$\pm$120.2$^\dagger$        & 23.9$\pm$8.2           & 190.8$\pm$65.8         & 46.4$\pm$33.8         & 0$^\ddagger$                 \\
                       & \checkmark  & \multicolumn{1}{l}{} & \multicolumn{1}{l}{} & \textbf{100} & 9.8$\pm$6.9$^\dagger$             & 13.0$\pm$3.8          & 182.2$\pm$43.9          & 44.5$\pm$29.1          & 24.7$\pm$22.1        \\
                       & \checkmark  & \checkmark                    & \multicolumn{1}{l}{} & \textbf{100} & \textbf{17.7$\pm$3.2}$^\dagger$   & \textbf{8.2$\pm$1.5}   & \textbf{141.6$\pm$12.9}   & 56.7$\pm$28.1          & 3.5$\pm$6.8         \\ \cline{6-6}
                       & \checkmark  & \multicolumn{1}{l}{} & \checkmark                    & \textbf{100} & 6.9$\pm$4.8             & 10.1$\pm$3.0           & 148.7$\pm$15.3          & 46.0$\pm$32.8          & 5.9$\pm$6.5          \\
                       & \checkmark & \checkmark                    & \checkmark                    & \textbf{100} & \textbf{7.4$\pm$5.5}    & \textbf{8.2$\pm$1.90} & 143.2$\pm$11.2            & \textbf{59.5$\pm$28.2} & \textbf{2.6$\pm$5.6} \\ \hline
\multirow{5}{*}{LCA}   & \multicolumn{3}{c}{Baseline}       & 0    & -212.7$\pm$106.1$\dagger$  & 16.8$\pm$0.4     & 286.4 6.1              & 42.22$\pm$30.1             & 0$\ddagger$    \\
                       & \checkmark  & \multicolumn{1}{l}{} & \multicolumn{1}{l}{} & \textbf{100} & 1.9$\pm$18.4$\dagger$               & 16.5$\pm$11.3             & 228.4$\pm$90.3             & 41.8$\pm$26.7            & 35.2$\pm$46.3             \\
                       & \checkmark  & \checkmark                    & \multicolumn{1}{l}{} & \textbf{100} & \textbf{6.7$\pm$6.5}$\dagger$       & 13.3$\pm$4.0               & 187.0$\pm$41.7             & 44.1$\pm$27.3            & 13.3$\pm$21.4             \\ \cline{6-6}
                       & \checkmark  & \multicolumn{1}{l}{} & \checkmark                    & \textbf{100} & -3.0$\pm$9.9               & 13.3$\pm$4.7              & 188.0$\pm$39.3    & 46.6$\pm$29.7  & 13.5$\pm$18.1             \\
                       & \checkmark  & \checkmark   & \checkmark    & \textbf{100} & \textbf{4.3$\pm$6.0}       & \textbf{11.0$\pm$2.3}      & \textbf{174.2$\pm$20.9}    & \textbf{52.2$\pm$31.1}    & \textbf{7.8$\pm$10.1}     \\ \hline
\multirow{5}{*}{BCA}   & \multicolumn{3}{c}{Baseline}                     & 15           & -122.3$\pm$87.7$^\dagger$         & 14.3$\pm$2.1           & 271.7$\pm$40.3          & 47.6$\pm$30.6         & 0$^\ddagger$                 \\
                       & \checkmark  & \multicolumn{1}{l}{} & \multicolumn{1}{l}{} & \textbf{100} & -3.4$\pm$12.0$^\dagger$           & 14.6$\pm$5.9           & 245.2$\pm$72.6          & 49.8$\pm$24.7          & 29.4$\pm$36.3       \\
                       & \checkmark  & \checkmark                    & \multicolumn{1}{l}{} & \textbf{100} & \textbf{4.8$\pm$2.6}$^\dagger$ & \textbf{10.4$\pm$0.8}         &    \textbf{184.5$\pm$4.9}       & 53.5$\pm$29.4          & 13.5$\pm$14.1        \\ \cline{6-6}
                       & \checkmark  & \multicolumn{1}{l}{} & \checkmark                    & \textbf{100} & -14.3$\pm$14.1         & 13.3$\pm$4.2           & 228.6$\pm$59.5         & 46.1$\pm$23.7          & 23.0$\pm$26.6        \\
                       & \checkmark  & \checkmark                   & \checkmark                    & \textbf{100} & \textbf{-8.3$\pm$8.3}  & 12.3$\pm$2.1   & 226.0$\pm$52.4 & \textbf{62.3$\pm$25.8} & \textbf{6.4$\pm$0.4} \\ \hline
\end{tabular}
}
\end{table*}

The quantitative experimental results from the tests are shown in Table~\ref{tab:vs}. Aside from the \textit{baseline}, all methods achieved autonomous guidewire navigation within the target areas of LSA, LCA and BCA with a 100\% success rate in fewer than maximum steps. This success is chiefly attributed to the strategy of employing the powerful pre-trained CNNs for feature extraction. Explicit observations and path navigation rewards can be considered as the explicit and implicit guidance for KVD agents in path navigation. Based on the pre-trained CNNs, explicit observations and path navigation rewards show improvements in episode rewards, episode lengths, movement distances, boundary distances, and retracement distances. Notably, \textit{baseline+PC+EO} achieves the highest episode lengths and movement distances in the LSA and BCA task, reaching the target with the fewest steps and shortest movement distance. Furthermore, path navigation rewards has significantly improved boundary distances, indicating that integrating pre-planned path and guidewire tip information into observations is beneficial for policy learning that tends towards the center of vessels. The integration of pre-trained CNNs, explicit observations, and path navigation rewards forms the core of the KVD algorithm, achieving the highest scores across numerous metrics. Specifically, in the more challenging LCA task, compared to the \textit{baseline+PC}, there was a reduction of 33.3\% in the average number of steps, a decrease of 23.7\% in the movement, and an increase of 24.9\% in the distance from the vascular boundary. Remarkably, the average distance for guidewire retraction was only 7.8 mm.

\subsection{Trajectory visualization}

\begin{figure}[t]
\centering
\includegraphics[scale=0.34]{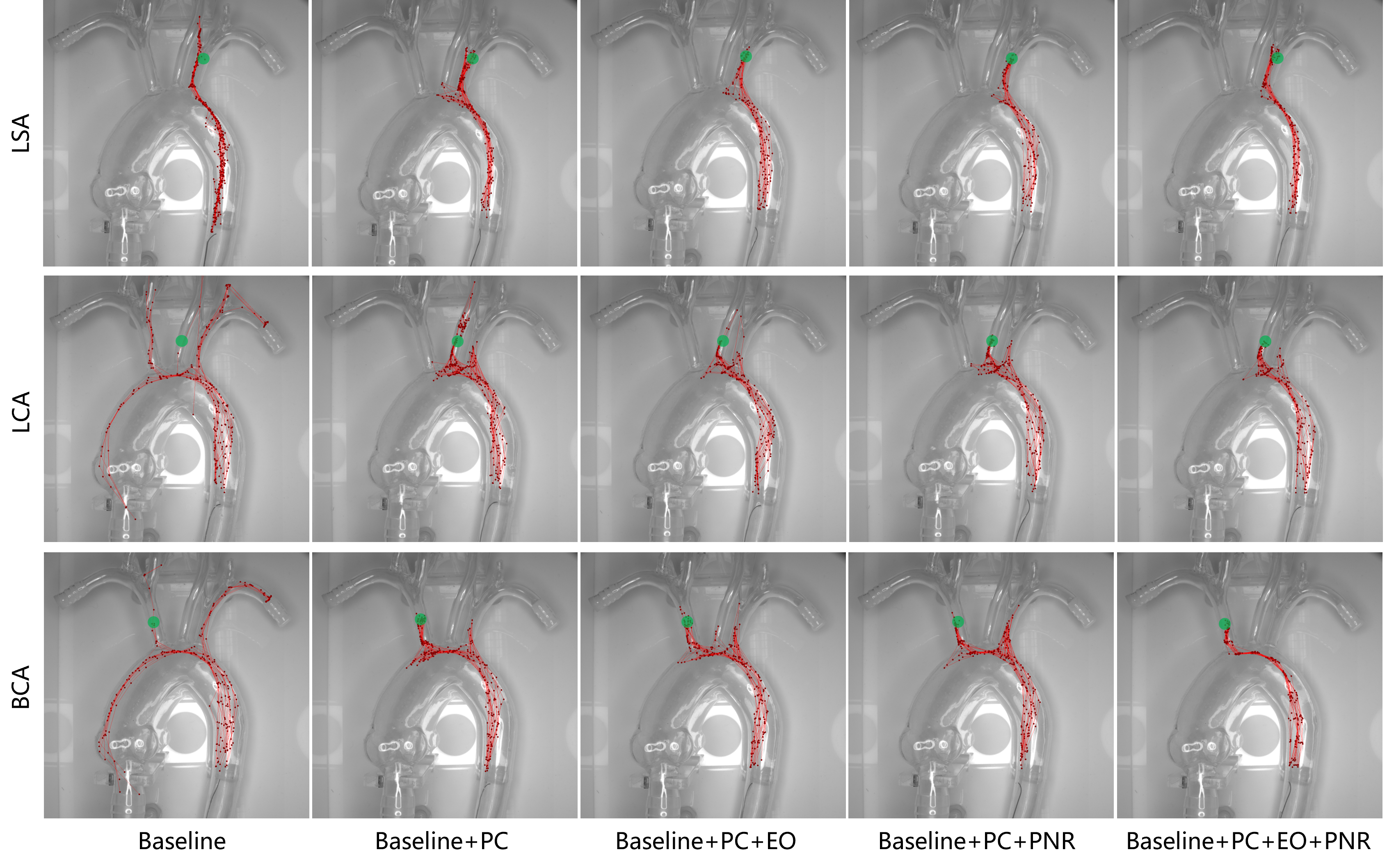}
\caption{Visualization of the path for autonomous guidewire navigation.} 
\label{fig:show}
\end{figure}

We visualized the guidewire movement trajectories during testing, as shown in Fig.~\ref{fig:show}. The performance of \textit{baseline} was poor, particularly in the LCA and BCA tasks, where most experiments resulted in the guidewire entering other vascular branches or ascending aorta, with the guidewire continuously moving forward without any action for retraction. This is consistent with the \textit{baseline}'s RD being 0, as listed in Table~\ref{tab:vs}, hence contributing to the lower success rate of guidewire navigation.  The \textit{baseline+PC} employs a pre-trained CNN to extract features and utilizes an MLP as the policy network. When the guidewire deviates from the planned path, the agent performs a retraction action. This is demonstrated in the LSA tasks, where the guidewire tip misses the LSA entrance and then returns, and in the BAC tasks, where the guidewire tip enters the LSA and then retracts back to the aortic arch. From the Fig.~\ref{fig:show}, it is evident that the path point distributions for  \textit{baseline+PC+EO} and  \textit{baseline+PC+PNR} are similar. Importantly, the combination of all components achieved the best results. The path points for \textit{baseline+PC+EO+PNR} are highly concentrated, with fewer retractions.

The proposed KVD method for vascular intervention robots achieved a 100\% success rate in wire delivery within an in vitro simulation model, demonstrating its potential application value. However, there are still significant challenges to applying this technology to actual clinical surgeries. First, although the method showed outstanding performance in tests following navigation training for specific targets, effectively applying the trained model to unknown target points or environments remains an urgent issue. This involves addressing the generalization capability of the autonomous guidewire navigation algorithm. Solving this problem will not only greatly enhance the method's practicality and flexibility but also serve as a key factor in its future adoption in clinical vascular interventions. Second, the method relies on extensive reinforcement learning training, which is impractical in clinical practice. While training in a virtual environment could mitigate this issue, constructing a virtual environment that accurately simulates the patient's vascular morphology and guidewire movement is a significant challenge. Such an environment must closely resemble real conditions and simulate complex hemodynamics and the physical properties of vascular walls. Therefore, future work should focus on improving model generalization, reducing training time, and developing more sophisticated virtual environment simulation technologies to advance autonomous guidewire navigation algorithms toward clinical application.

\subsection{Efficiency Comparison on a Commercial Robot}

\begin{figure}[ht]
\centering
\includegraphics[scale=0.36]{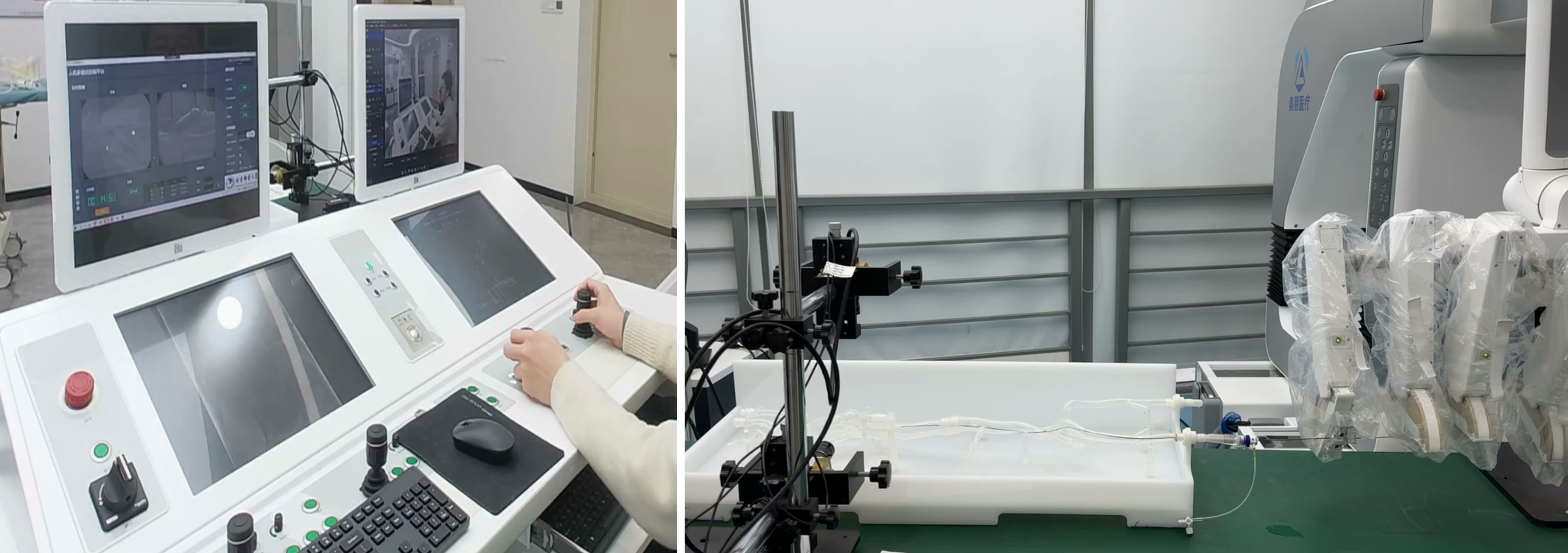}
\caption{ALLVAS interventional surgical robot experimental platform: The left image shows the master-slave control and real-time display platform, while the right image shows the robotic arms of the vascular intervention system.} 
\label{fig:aopeng}
\end{figure}

To validate whether KVD can effectively reduce surgery time compared to human-machine collaborative operations, the algorithm was deployed on the ALLVAS vascular intervention robot developed by Shanghai Aopeng Medical Technology Co., Ltd. The ALLVAS consists of two operating arms and a control console. By synchronizing the gripper mechanisms at the end of the two arms, the robot can rotate the guidewire left/right and advance/withdraw it (as shown in the right image of Figure ~\ref{fig:aopeng}), until the guidewire reaches the target vascular region. As shown in the left image of Figure ~\ref{fig:aopeng}, the Master-Slave Control (MSC) console is equipped with two joysticks and four knobs. The two joysticks are used to control the push-pull and rotation of the guidewire through the robotic arms, while the knobs adjust the robot's operating speed and start/stop functions. Commands from both manual operators and the agent were transmitted to the console via TCP/IP protocol to control the ALLVAS. Professionally trained operators monitored the real-time front and side views of the vascular simulation model displayed on the console to perform MSC for guidewire navigation. In the efficiency comparison between autonomous and manual guidewire navigation, the tip of the guidewire was initially placed inside the descending aorta, and the target vessels for the experiments were the LSA and BCA. Each task was performed 5 times using both autonomous guidewire navigation and operator-controlled master-slave navigation, and the time to guide the wire to the LSA and BCA was recorded for both modes.

\begin{figure}[ht]
\centering
\includegraphics[scale=0.30]{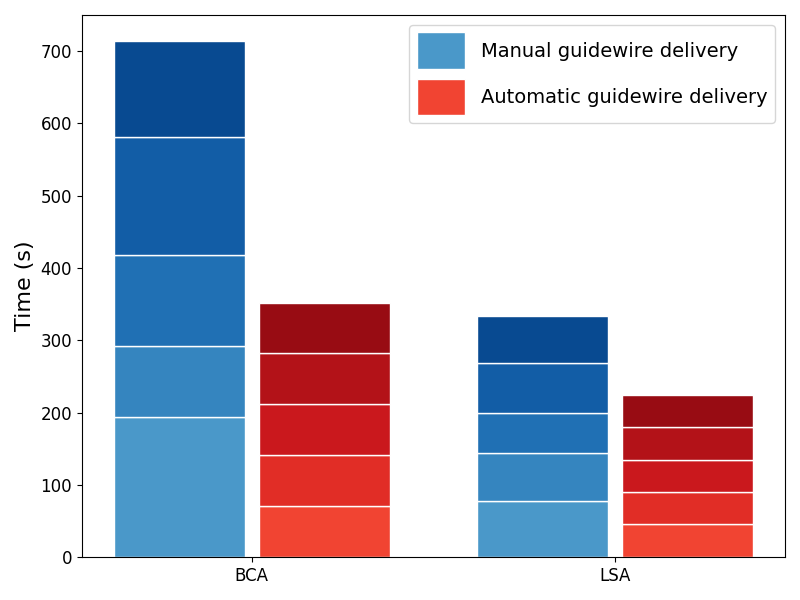}
\caption{Comparison of guidewire navigation time between autonomous operation and human-machine collaborative operations} 
\label{fig:timevs}
\end{figure}

The experimental results, as shown in Figure~\ref{fig:timevs}, indicate that the blue bar chart represents the time taken for guidewire navigation using the MSC mode, while the red bar chart represents the time taken using KVD. In the BCA task, the average time for guidewire navigation under the MSC mode was 142.8 s, while the autonomous mode averaged 70.4 s. For the LCA task, the average times were 66.9 s for the master-slave mode and 44.9 s for the autonomous mode. Compared to the MSC mode, the autonomous guidewire navigation mode reduced the procedure time by 50.7\% and 32.9\% for the BCA and LCA tasks, respectively, significantly improving navigation efficiency. Notably, in the more complex BCA task, the autonomous guidewire navigation mode demonstrated superior performance. It's also worth mentioning that in the MSC mode, operators performed guidewire navigation by observing two orthogonal views of the vascular model. However, during real aortic interventions, DSA imaging equipment often uses a single C-arm design, meaning that surgeons typically rely on a single view of aortic angiography to perform the procedure. This results in less imaging information for guidance, potentially leading to longer navigation times in the MSC mode. Additionally, the bar chart for the BCA task in Figure~\ref{fig:timevs} shows significant variation between trials in the MSC mode, indicating that this mode produces inconsistent navigation performance. In contrast, the autonomous guidewire navigation mode demonstrated greater consistency.


\section{Conclusions}
In this work, we propose a RL-based KVD method for robot-assisted EI. This method employs the proposed BDA-star, guiding toward the vascular center, to plan the paths for KVD, establishing a RL environment comprised of observations that integrate the guidewire tip and planned path, along with a path navigation reward function. Notably, we address the challenge of slow convergence during policy learning by employing a network that combines a pre-trained CNN with a MLP. Experiments on the aortic simulation KVD platform demonstrated a 100\% success rate in guidewire navigation for the LSA, LCA and BCA, along with reduced movement and retraction distances, and the trajectories converged towards the central axis of the vessels, revealing its potential is clinical practice.

\bibliographystyle{IEEEtran}
\bibliography{refer}\ 

\end{document}